\documentclass[runningheads]{llncs}

\usepackage[T1]{fontenc}
\usepackage{graphicx}

\usepackage{xcolor}

\usepackage{amsmath}
\usepackage{amssymb}
\usepackage{latexsym}
\usepackage{booktabs}
\usepackage{enumitem}
\usepackage{color}
\usepackage{bm}
\usepackage{multirow}
\usepackage{enumitem}
\usepackage{float}
\usepackage{svg}
\usepackage{titletoc}
\usepackage{multicol}
\usepackage{makecell}
\usepackage{xurl}
\usepackage{upgreek}
\usepackage{hyperref}

\usepackage{array}
\newcolumntype{L}[1]{>{\raggedright\let\newline\\\arraybackslash\hspace{0pt}}m{#1}}
\newcolumntype{C}[1]{>{\centering\let\newline\\\arraybackslash\hspace{0pt}}m{#1}}
\newcolumntype{R}[1]{>{\raggedleft\let\newline\\\arraybackslash\hspace{0pt}}m{#1}}

\usepackage{stackengine}

\newcolumntype{P}[1]{>{\centering\arraybackslash}p{#1}}

\begin{document}
\title{Emerging NeoHebbian Dynamics in Forward-Forward Learning: Implications for Neuromorphic Computing}
%
%
\author{Erik B. Terres-Escudero\inst{1}\orcidID{0009-0003-9781-7657} \and
Javier Del Ser\inst{2,3}\orcidID{0000-0002-1260-9775} \and
Pablo Garc\'ia-Bringas\inst{1}\orcidID{0000-0003-3594-9534}}

\authorrunning{E. B. Terres-Escudero, J. {Del Ser}, P. {Garc\'ia-Bringas}}
\titlerunning{NeoHebbian Dynamics in Forward-Forward Learning}

\institute{DeustoTech, University of Deusto, 48014 Bilbao, Spain \\ \email{\{e.terres, pablo.garcia.bringas\}@deusto.es} \and
TECNALIA, Basque Research \& Technology Alliance (BRTA),\\48160 Derio, Spain. \email{javier.delser@tecnalia.com}\\
\and
University of the Basque Country (UPV/EHU), 48013 Bilbao, Spain}

\maketitle              
\begin{abstract}

Advances in neural computation have predominantly relied on the gradient backpropagation algorithm (BP). However, the recent shift towards non-stationary data modeling has highlighted the limitations of this heuristic, exposing that its adaptation capabilities are far from those seen in biological brains. Unlike BP, where weight updates are computed through a reverse error propagation path, Hebbian learning dynamics provide synaptic updates using only information within the layer itself. This has spurred interest in biologically plausible learning algorithms, hypothesized to overcome BP's shortcomings. In this context, Hinton recently introduced the Forward-Forward Algorithm (FFA), which employs local learning rules for each layer and has empirically proven its efficacy in multiple data modeling tasks. In this work we argue that when employing a squared Euclidean norm as a goodness function driving the local learning, the resulting FFA is equivalent to a neo-Hebbian Learning Rule. To verify this result, we compare the training behavior of FFA in analog networks with its Hebbian adaptation in spiking neural networks. Our experiments demonstrate that both versions of FFA produce similar accuracy and latent distributions. The findings herein reported provide empirical evidence linking biological learning rules with currently used training algorithms, thus paving the way towards extrapolating the positive outcomes from FFA to Hebbian learning rules. Simultaneously, our results imply that analog networks trained under FFA could be directly applied to neuromorphic computing, leading to reduced energy usage and increased computational speed.

\end{abstract}

\section{Introduction}

For many decades the Artificial Intelligence (AI) community has drawn inspiration from neuroscience to develop novel neural architectures and training algorithms. However, since the introduction of the backpropagation (BP) algorithm, the field has experienced a drastic shift, with most modern neural computation approaches extending from the principles introduced by BP. While this algorithm has proven effective in solving a wide range of tasks, several shortcomings (e.g. Catastrophic Forgetting \cite{goodfellow2013empirical}) render BP inefficient when dealing with non-stationary data distributions. To address this limitation, the NeuroAI movement \cite{zador2023catalyzing} has recently gained momentum, with the premise that a better understanding of neural systems will lead to the development of optimal learning algorithms that exhibit more adaptable learning dynamics. 

Among the biologically inspired methods contributed to date, forward-only algorithms stand as efficient alternatives to BP achieving competitive results in multiple conventional learning tasks. These algorithms replace the backward propagation path with multiple local learning rules, thereby addressing key biological implausibilities such as the weight symmetry problem \cite{ororbia2023brain}. In this field, this work focuses on the recently introduced Forward-Forward Algorithm (FFA) \cite{hinton2022forward}, which has demonstrated competitive performance compared to BP. However, despite being motivated by biological constraints, only one work has studied the relation between FFA and its underlying biological plausibility \cite{ororbia2023contrastive}. In this work, Ororbia developed a spiking alternative for the conventional formulation of FFA, replacing the error-based weight updates with a Hebbian alternative, achieving competitive results compared to its analog counterpart.

This paper takes an step beyond the results reported in \cite{ororbia2023contrastive} by analyzing the relationship between a generalized formulation of FFA and the dynamics presented by Hebbian learning. We show that by employing a squared Euclidean norm as a goodness function, the resulting learning rule is equivalent to a modulated Hebbian learning rule. To empirically verify our claim, we train several models using the conventional formulation of FFA and its Hebbian adaptation, subsequently comparing their behavioral similarities in the generated latent space. As a byproduct of this relationship, we elaborate on the potential of FFA to become an emergent framework for developing Hebbian learning solutions leveraging the speed and energy advantages of neuromorphic systems, exposing a promising synergy between both research areas. 

The rest of the paper is structured as follows: Section \ref{sec:related_work} offers a brief introduction to FFA and frames the contribution of this paper within the existing literature. Next, Section \ref{sec:relation_1} details the main theoretical result of the paper, addressing the relationship between FFA and Hebbian theory. Section \ref{sec:exp_setup} formulates research questions and presents the setup used to answer them with evidence. Section \ref{sec:result} presents and discusses the results from the aforementioned experiments. Implications in biological learning and neuromorphic computing are examined in Section \ref{sec:impact}. Finally, Section \ref{sec:conclusion} offers a conclusion of the paper.

\section{Related Work}
\label{sec:related_work}

Forward-only approaches represent an alternative family of learning algorithms in which the conventional backward pass that lies at the core of BP is replaced by a secondary forward pass to estimate the network's error. This procedure allows the model to update its trainable parameters (\emph{weights}) without requiring information about the error of subsequent layers, thereby eliminating the update-lock and weight-transport issues \cite{ororbia2023brain}. Among the algorithms in this family, the FFA \cite{hinton2022forward} has recently proven to be a competitive alternative to BP, exhibiting biologically plausible latent representations characterized by sparsity \cite{yang2023theory} and high neural specialization \cite{ororbia2023predictive}. 

This algorithm operates through a contrastive process, in which the model is trained to distinguish between real and synthetic images, referred to as positive samples $D_{\oplus}$ and negative samples $D_{\ominus}$, respectively. To classify the input, networks resorting to the FFA employ layer-specific loss functions that drive the weight updates using solely information from the latent activity vector $\bm{\ell} \in \mathbb{L}$, where $\mathbb{L}$ is the latent space of the respective layer. In his original approach, Hinton proposed the definition of a goodness function $G : \mathbb{L} \rightarrow \mathbb{R}$ to measure the observed fitness of the latent form belonging to a sample from the positive data distribution. Due to its simple derivative, the squared Euclidean norm was initially proposed. Under this formulation, the latent activity vector $\bm{\ell}_{\oplus}$ of an input sample $\mathbf{x}_{\oplus} \in D_{\oplus}$ would be considered optimal if \( G(\bm{\ell}_{\oplus}) \) obtains high distances from the origin, while the activity vector \( \bm{\ell}_{\ominus} \) of a negative sample \(\mathbf{x}_{\ominus} \in D_{\ominus} \) would be deemed optimal only if \( G(\bm{\ell}_{\ominus}) \) minimized the goodness function, thereby neighboring the origin of the latent space. To avoid relying on a regression-based formulation, a probability function \( P : \mathbb{R} \rightarrow [0, 1] \) is employed to map goodness scores into a probability distribution, measuring the probability of an input sample from been extracted from the positive data distribution. The earliest formulation of this function was given by the sigmoid function \( P_{\sigma}(\psi; \theta, \alpha) = \texttt{sigmoid}(\alpha (\psi - \theta)) \), where \( \alpha \) is a scaling factor, $\texttt{sigmoid}(\cdot)$ is the Sigmoid function, and \( \theta \) is a threshold value to shift the probability distribution \cite{hinton2022forward}. Ideally, an optimally trained model under this formulation would have negative samples $\mathbf{x}_{\ominus} \in D_{\ominus}$ with low goodness scores achieve low probability scores, $P_{\sigma}(G(\mathbf{x}_{\ominus}); \theta, \alpha) \approx 0$. Conversely, positive samples $\mathbf{x}_{\oplus} \in D_{\oplus}$ with high goodness scores would lead to high values of the probability distribution, $P_{\sigma}(G(\mathbf{x}_{\oplus}); \theta, \alpha) \approx 1$. By using this probability function, the model is trained to minimize a Binary Cross-Entropy loss function through conventional gradient descent algorithms.

Although FFA was shown to achieves competitive results compared to BP on several datasets, the original formulation of this learning algorithm is subject to  shortcomings that can reduce its accuracy. A primary limitation was identified by Lee et al. in \cite{lee2023symba}, who argued that the original formulation suffers from a gradient vanishing problem during training on negative samples, resulting in uneven learning dynamics and suboptimal performance. To address this issue, an alternative formulation of FFA can be achieved by incorporating a polarity division into each layer. This division assigns a polarity to each neuron, either positive or negative, so that the subsequent training objective shifts from maximizing the positive goodness to the maximization of the goodness score of each neuron with respect to its polarity type. By making negative samples follow an equivalent objective to positive inputs, the gradient achieves an even distribution, thereby solving the vanishing problem and achieving a more stable learning process. An alternative probability function, denoted as the symmetric probability $\textup{P}_{\text{Sym}}$, can be defined as the ratio between the goodness of the positive neurons and the overall activity of the layer.

Under this learning framework, the behavior of any layer is driven by the selection of three fundamental choices: the activation function, the goodness function, and the probability function. Therefore, given a learning rate $\eta$, and being $\bm{a}_{\oplus}^k$ the pre-activation vectors of the latent activities $\bm{\ell}_{\oplus}^k$, the update dynamics of any FFA model are given by:
\begin{equation}
    \label{eq:weight_update}
    \frac{\partial L_{\textup{CE}}}{\partial w_{ij}}= - \frac{1}{K}\sum_{k=1}^K \frac{\eta}{\Upsilon(G(\bm{\ell}_{\oplus}^k), G(\bm{\ell}_{\ominus}^k))}\frac{\partial \Upsilon}{\partial G} \frac{\partial G}{\partial \ell_{\oplus,j}^k} \frac{\partial \ell_{\oplus,j}^k}{\partial a_{\oplus, j}^k} x_{i}^k,
\end{equation}
where $L_{\textup{CE}}$ denotes the cross-entropy function; $\eta\in\mathbb{R}^+$ is the learning rate; $k$ refers to sample index within the $K$-sized batch; $x_i^k$ is the $i$-th input of the $k$-th data instance $\mathbf{x}^k$; $\Upsilon(G_\oplus,G_\ominus)$ refers to any probability function that combines the activations of both positive and negative instances associated to instance $k$, e.g. $\Upsilon\left(G(\bm{\ell}_\oplus),G(\bm{\ell}_\ominus)\right)=G(\bm{\ell}_\oplus)/(G(\bm{\ell}_\oplus)+G(\bm{\ell}_\ominus))$
for the symmetric probability function and $\Upsilon\left(G(\bm{\ell}_\oplus),G(\bm{\ell}_\ominus)\right)=\texttt{sigmoid}(\alpha(G(\bm{\ell}_\oplus)-\theta))$ for the sigmoid probability function; and $j$ refers to neuron index within the layer, such that $\ell_{\oplus,j}^k$ stands for the positive activation of instance $k$ at the $j$-th output neuron. A similar expression holds for the negative instances by replacing $\ell_{\oplus,j}^k$ with $\ell_{\ominus,j}^k$.

\paragraph{Contribution:} Although the primary motivation for the development of FFA has been to create biologically plausible alternatives to BP, there has been little research exploring this relationship. To date, only a limited number of studies have addressed the biological underpinnings of this algorithm \cite{ororbia2023contrastive,ororbia2023predictive}. Among them, the most in-depth analysis was conducted by Ororbia, who adapted FFA to operate on spiking models by replacing analog optimization methods with a modulated Hebbian learning rule, performing on par with its original implementation \cite{ororbia2023contrastive}. Our work builds on his results to demonstrate that FFA, when using Euclidean goodness functions, naturally produces Hebbian update dynamics, and thus can seamlessly be trained in spiking neural networks.

\section{Relationship between neoHebbian Learning and the Forward-Forward Algorithm}
\label{sec:relation_1}

Hebbian learning rules have been among the earliest mechanisms studied in terms of learning and memory formation in the brain. Neurons updated by this rule see their synapses strengthen whenever pre-synaptic and post-synaptic spikes occur closely within short time frames. However, their traditional formulation does not account for complex learning dynamics mainly driven by neuromodulatory activity. To address this limitation, neoHebbian learning rules modify the original formulation by introducing an additional factor that regulates the strength and polarity of the update \cite{gerstner2018eligibility,lisman2011neohebbian}. Formally, given a pre-synaptic neuron \(x_i\), and a post-synaptic neuron \(y_i\), a neoHebbian update of the weight  \( w_{ij} \) connecting them is given by:
\begin{equation}
    \label{eq:neohebbian_eq}
    \frac{\partial w_{ij}}{\partial t} \doteq \Delta_t w_{ij} = M(t) g(y_j) x_i,
\end{equation}
where \( M(t) \) represents the so-called \emph{third factor} modulating the update at time $t$, and \( g(\cdot) \) is an aggregation operator of the spiking activity $y_j$ of neuron $j$. The third factor is a signal that modulates the plasticity of synaptic connections, influencing the learning speed and the direction of change in synaptic weights. Its form depends on the learning task under consideration, e.g. it can be a global scalar related to reward in Reinforcement Learning tasks, or a neuron-specific signal related to the network output error in supervised learning.

As mentioned previously, Hinton originally proposed the squared Euclidean norm as the goodness function for his proposed FFA formulation. This architectural choice was primarily driven by the simple derivative this function generates, which yields a straightforward \(\nabla G(\bm{\ell}) = 2\bm{\ell}\). By substituting this value into the goodness derivative in Equation \eqref{eq:weight_update}, it becomes evident that the resulting expression contains the same factors as the neoHebbian update rule in Expression \eqref{eq:neohebbian_eq}. For instance, when using the aforementioned goodness function and a batch size equal to 1 -- index $k$ can be dropped from Expression \eqref{eq:weight_update} -- the weight update function is given by:
\begin{equation}
    \label{eq:weight_update_post}
    \frac{\partial L_{\textup{CE}}}{\partial w_{ij}}= - \frac{2 \eta}{\Upsilon(G(\bm{\ell}_{\oplus}), G(\bm{\ell}_{\ominus}))}\frac{\partial  \Upsilon}{\partial G} \frac{\partial \ell_{\oplus,j}}{\partial a_{\oplus, j}} \ell_{\oplus,j}  x_{i}.
\end{equation}

Within this expression, the pre-synaptic and post-synaptic multiplication of the traditional Hebbian learning rule can be related via \( y_j \cdot x_i \sim \ell_{\oplus,j} \cdot x_i \) (with an equivalent formulation for the negative instance of \( \ell_{\ominus} \)). Correspondingly, we can link the additional multiplicative terms in Expression \eqref{eq:weight_update_post}, which comprise the remaining derivatives of the original error formula, with the neoHebbian modulatory factor \( M(t) \) defined as:
\begin{equation}
M(t) = \frac{2 \eta }{\Upsilon(G(\bm{\ell}_{\oplus}), G(\bm{\ell}_{\ominus}))}\frac{\partial  \Upsilon}{\partial G} \frac{\partial \ell_{\oplus,j}}{\partial a_{\oplus, j}}.
\end{equation}
where $\ell_{\oplus,j}$ denotes the output activity of neuron $j$. When using a spiking implementation, it is important to note that $\ell_{\oplus,j}\equiv \ell_{\oplus,j}(t)$, i.e. the activity of the neuron becomes a function of time. We hereafter omit this explicit dependence on time when understood from the context.

Unfortunately, the non-differentiable dynamics of spiking neurons pose a challenge when computing the derivative of the post-activation value \(\ell_{\oplus,j}\). To address this limitation, we advocate for approximating this value using an output trace, which provides a smooth approximation of the spiking dynamics. Analogously, we rely on the value of this trace for computing the probability and goodness functions, as it helps the weight updates avoid abrupt changes and increases the stability of the learning phase.

By employing this construction, we can verify that all configurations implementing FFA-like learning heuristics exhibit neo-Hebbian behaviors, rendering them as biologically plausible alternative learning rules for both analog and spiking networks. To exemplify this, we present the weight update expressions of two distinct configurations of FFA: one employing its conventional form with a sigmoid probability $P_\sigma(\cdot)$; and the other using the symmetric implementation $P_{\textup{Sym}}(\cdot)$ with split layers. Since these expressions are computed in a spiking domain, we extend the previously introduced notation to its equivalent spiking form. In this scenario, \(\ell_{\oplus,j}\) represents the output trace from a positive neuron \(j\) employing ReLU-like dynamics. The characteristic update functions from Equation \(\eqref{eq:weight_update}\) of the sigmoid and symmetric probabilities are given by: 
\begin{align}
\label{eq:sigmoid_update}
\text{Sigmoid probability:}  \quad  & \displaystyle\Delta w_{ij} = (1-P_{\sigma}(G(\bm{\ell}))) \ell_{j} x_i, \\ \label{eq:symmetric_update}\text{Symmetric probability:} \quad & \displaystyle\Delta w_{ij} = \frac{1 - P_{\text{Sym}}(G(\bm{\ell}_{\oplus}),G(\bm{\ell}_{\ominus}))}{G(\bm{\ell}_{\oplus})}\ell_{\oplus,j} x_i.  
\end{align}

This learning rule formulation can be extended to incorporate additional biological mechanisms in order to enhance the learning dynamics. One such heuristic can be obtained mimicking conventional optimizers (e.g., ADAM or RMSProp), with the goal of modulating the update so as to capture the momentum generated through the learning process. An example found in biology is given by eligibility traces \cite{gerstner2018eligibility}, where weight updates are gradually introduced into the network as time passes instead of being directly computed. This approach provides additional stability to the training process, which can be crucial in scenarios where batch sizes are small (e.g. online learning). Formally, a trace \( e_{ij} \) captures the original impulse of the weight updates, which are computed through the previously defined loss function. Subsequently, this trace is then gradually aggregated into the weight \( w_{ij} \), while a decaying factor diminishes the cumulative importance of older weight updates stored in \( e_{ij} \) over time, favoring the momentum obtained from recent updates. Formally, the trace can be defined as:
\begin{align}
    \Delta e_{ij} &=  (1-\tau_{e})\left(\frac{\partial L_{\textup{CE}}}{\partial w_{ij}} - e_{ij}\right), \label{eq:elegibility_trace_equation_1}\\
    \Delta w_{ij} &= \eta \cdot e_{ij}, \label{eq:elegibility_trace_equation_2}
\end{align}
where $\tau_{e}\in\mathbb{R}[0,1]$ represents the decay factor of the eligibility trace, and $\eta \in \mathbb{R}^+$ denotes the learning rate of the update rule.

\section{Experimental Setup}
\label{sec:exp_setup}

In the previous section we presented the theoretical equivalence between FFA and Hebbian Learning, exposing that the modulation factor can be linked to the derivatives of the activation and probability functions in the FFA's update rule. To formally assess these results, we propose a set of experiments designed to answer the following \emph{Research Questions} (RQ):
\begin{itemize}[leftmargin=*]
    \item {\textbf{RQ1}}: Do biological implementations of FFA relying on Hebbian learning rules perform competitively when compared to analog FFA implementations?
    \item {\textbf{RQ2}}: Do learning mechanics of Hebbian FFA result in equivalent latent spaces to the ones obtained in the analog implementation of FFA?
\end{itemize}

To address RQ1, we assessed the accuracy levels obtained by training different spiking neural configurations on the MNIST dataset. These configurations involve primary functions detailed in Equation \eqref{eq:weight_update}: the probability function, which can be either the sigmoid probability \(P_{\sigma}\) or the symmetric probability \(P_{\text{Sym}}\); and the output trace, which can be regarded as the activation function in spiking networks. The output trace is chosen among the LI trace, ReLU trace, or Hard-LI trace, which are mathematically defined and explained in detail in Appendix \ref{ap:out_trace}. We conducted an initial exploration of the hyper-parameter space of spiking models configured with different values of the learning rate $\eta\in\{0.001, 0.01, 0.1, 1, 10\}$ and the decay factor of the eligibility trace $\tau_{e}\in\{0.999, 0.99, 0.9\}$. Based on these results, we selected the configurations achieving the highest accuracy in just one epoch, used an extended training of 10 epochs for the reported results. To further explore the performance of Hebbian FFA under biologically plausible scenarios, we compared its performance in batch scenarios, using a batch size of $K=50$ samples, and online scenarios (i.e. $K=1$). To validate whether these configurations perform on par against their analog counterpart, we trained an analog neural network model with an equivalent architecture using the same probability functions. All training experiments employed a supervised learning approach defined in the original paper by Hinton \cite{hinton2022forward}. This approach involves embedding a label into the input data, with positive samples using their respective label and negative samples having an incorrect label assigned to create the contrastive process. To embed the labels, we employ the method described in Lee et al., in which each label is replaced with a binary vector that can be appended to the end of the image \cite{lee2023symba}. These binary vectors were created by uniform sampling with \( p = 0.3 \).

Although the inclusion of multiple layers are often associated to enhanced representational power of the model and potentially increased accuracy \cite{hinton2022forward}, the objective of this experimentation does not prioritize achieving high-performing models. Therefore, considering that the local training mechanisms of FFA can demonstrate effective learning capabilities even with a single layer, we employ networks consisting of one layer with 200 neurons to reduce computational costs, especially in online learning tasks. The spiking layers use a LIF (\emph{Leaky Integrate-and-Fire}) neural model with a decay factor of $0.85$. All spiking neural models employ the eligibility trace described in Equations \eqref{eq:elegibility_trace_equation_1} and \eqref{eq:elegibility_trace_equation_2} to smooth weight updates during training. In contrast, analog neural models employ a ReLU activation function, which closely resembles the dynamics of the output traces used in their spiking counterparts. The input data are encoded into spiking activity through a rate-based encoding scheme with a scaling factor of $0.25$. Models are trained using 24 time steps per sample, with Hebbian weight updates active only during the last $9$ time steps. Finally, the analog networks are trained using an ADAM optimizer with a learning rate of $0.01$.

To provide an answer to RQ2, we studied two key properties that characterize latent space of FFA models: latent sparsity and class separability. To examine the first property, we generated the latent vector of all instances within the test datasets, and subsequently computed the \emph{Hoyer Index} of each representative latent space. This index assigns an score in the range $\mathbb{R}[0,1]$ to a given latent vector $\bm{\ell}\in\mathbb{R}^n$, where $0$ represents a uniformly distributed activity and $1$ a vector with only one active unit. The Hoyer Index $\textup{HI} : \mathbb{R}^{n} \rightarrow \mathbb{R}[0,1]$ is given by:
\begin{equation}
\textup{HI}(\bm{\ell}) = \frac{\sqrt{n} - \|\bm{\ell}\|_1/\|\bm{\ell}\|_2}{\sqrt{n}-1}.
\end{equation}

Conversely, to examine the embedding representation of the different classes and verify their cohesiveness, we rely on two methods. First, we provide a qualitative analysis of the projections obtained from T-SNE, comparing the geometrical properties that emerge from this projection. Additionally, we rely on the \emph{Separability Index} \cite{thornton1998separability} to quantitatively measure the distance between the different class clusters. This index returns the average ratio of neighbors of a sample that have the same assigned label. Scores closer to \(1\) imply more separated spaces. The separability index \(\textup{SI}(\bm{\mathcal{L}}, \bm{\mathcal{C}})\) of a set of latent vectors \(\bm{\mathcal{L}} = \{\bm{\ell}_1, \bm{\ell}_2, \ldots,\bm{\ell}_Q\}\) with their respective class labels \(\bm{\mathcal{C}} = \{c_1, c_2, \ldots,c_Q\}\) is computed as:
\begin{equation}
\textup{SI}(\bm{\mathcal{L}}, \bm{\mathcal{C}}) = \frac{1}{Q\cdot K_{nn}}\sum_{q=1}^{Q} \sum_{k_{nn}=1}^{K_{nn}} \delta(c_q, \textup{cKNN}_{k_{nn}}(\bm{\ell}_q)),
\end{equation} 
where $Q$ denotes the number of samples in $\bm{\mathcal{L}}$; \(K_{nn}\) is the number of neighbors compared, which has been set to \(5\) for our experiments; and \(\textup{cKNN}_{k_{nn}}(\bm{\ell}_q)\) represents the class corresponding to the \(k_{nn}\)-th euclidean nearest neighbor of \(\bm{\ell}_q\in\bm{\mathcal{L}}\). Additionally, \(\delta(a,b)\) represents the Kronecker delta, which outputs \(0\) if \(a \neq b\) and \(1\) otherwise.

For the sake of reproducibility and to stimulate follow-up studies, the scripts and results discussed in what follows are available in \url{https://github.com/erikberter/Hebbian_FFA}.

\section{Results and Discussion} \label{sec:result}

\paragraph{RQ1:} 

Results from the experiments corresponding to this first research question are presented in Table \ref{tab:accuracies}, which summarizes the test accuracy on the MNIST dataset achieved by an analog FFA, a Hebbian FFA and an \emph{online} version of the Hebbian FFA, the latter configured with a single-instance batch size ($K=1$). Results are reported for the two probability functions $P_\sigma$ and $P_{\textup{Sym}}(\cdot)$ leading to the two definitions of $\Upsilon(\cdot,\cdot)$ in the FFA update rule of Expressions \eqref{eq:weight_update} (analog FFA) and \eqref{eq:sigmoid_update}-\eqref{eq:symmetric_update} (Hebbian FFA). Among all these results, the Hebbian implementation of FFA achieves competitive accuracy levels compared to its analog counterpart. In addition, the symmetric probability function \( P_{\textup{Sym}} \) systematically obtains better performance across all experiments than the sigmoid probability function. Remarkably, the analog version of FFA dominates the benchmark when configured with both probability functions; however, this enhanced performance can be potentially attributed to the increased number of hyper-parameters of LIF neurons in its non-analog counterparts (e.g., neuronal decay, input strength), which require additional computational power to reach optimal scores. This effect could be mitigated through a more extensive hyper-parameter tuning process. All in all, the relative small gaps reported in the table confirm that the analog and Hebbian implementations of FFA yield near-performing neural networks, positively answering RQ1. Furthermore, the comparison between batched and online implementations of Hebbian FFA reveal even closer accuracy levels, demonstrating that online, biologically driven implementations can achieve competitive performance without significant accuracy drops.
\begin{table}[t!]
    \centering
    \caption{Test accuracy on the MNIST dataset achieved by the Analog, Hebbian, and Online Hebbian models using the sigmoid $P_\sigma$ and symmetric $P_{\textup{Sym}}$ probabilities.}
    \label{tab:accuracies}
    \begin{tabular}{lC{2cm}C{2cm}C{3.2cm}}
         \toprule[1.2pt]
         \midrule
         Probability & Analog FFA &  Hebbian FFA & Online Hebbian FFA \\
         \midrule
         Sigmoid ($P_{\sigma}$) &   89.74\% & 86.21\% & 85.11\% \\
         Symmetric ($P_{\text{Sym}}$) & 95.10\% & 92.72\% &94.36\% \\
         \bottomrule
         \bottomrule[1.2pt]
\end{tabular}
\vspace{-3mm}
\end{table}
\begin{table}[h]
\vspace{-2mm}
    \centering
    \caption{Accuracy of the models on the MNIST dataset under the Sigmoid and Symmetric probabilities for the Hebbian, and Online Hebbian models broken down by output trace.}
    \label{tab:model_trace}
    \begin{tabular}{lC{1.35cm}C{1.35cm}C{1.35cm}C{.5cm}C{1.35cm}C{1.35cm}C{1.35cm}}
    \toprule[1.2pt]
    \midrule
    & \multicolumn{3}{c}{Hebbian FFA} & & \multicolumn{3}{c}{Online Hebbian FFA} \\
    \cmidrule{2-4} \cmidrule{6-8}
            Probability                   & LI   & ReLU  & Hard-LI  & & LI   & ReLU  & Hard-LI \\
            \midrule
    Sigmoid ($P_{\sigma}$)    & 79.35\%  & 81.87\%  & 63.67\% & & 85.07\% &  83.92\% &  74.11\%\\
    Symmetric ($P_{\text{Sym}}$)& 88.95\%  & 86.87\% & 85.54\% & & 87.13\% &  82.15\% &  85.55\%\\
    \bottomrule
    \bottomrule[1.2pt]
    \end{tabular}
    \vspace{-5mm}
\end{table}

A further inspection of the results for RQ1 can be made by examining how the performance varies with the trace of the outputs in use. The results of this second set of experiments for RQ1 are presented in Table \ref{tab:model_trace}, suggesting that higher accuracy scores can be achieved by employing a ReLU-like trace. These findings support the hypothesis in Section \ref{sec:relation_1}, which posited that output traces with derivative dynamics closer to their original functions would yield more accurate results. In contrast, the Hard-LI trace achieves the lowest average accuracy: this degraded performance is likely due to sporadic spikes pushing the trace abruptly to high activity states, resulting in imprecise and noisy outputs. Consistently with the results from Table \ref{tab:accuracies}, the online implementations of FFA perform closely to their batched counterparts.

\begin{figure}[h]
\vspace{-2mm}
    \centering
    \includegraphics[width=0.9\textwidth]{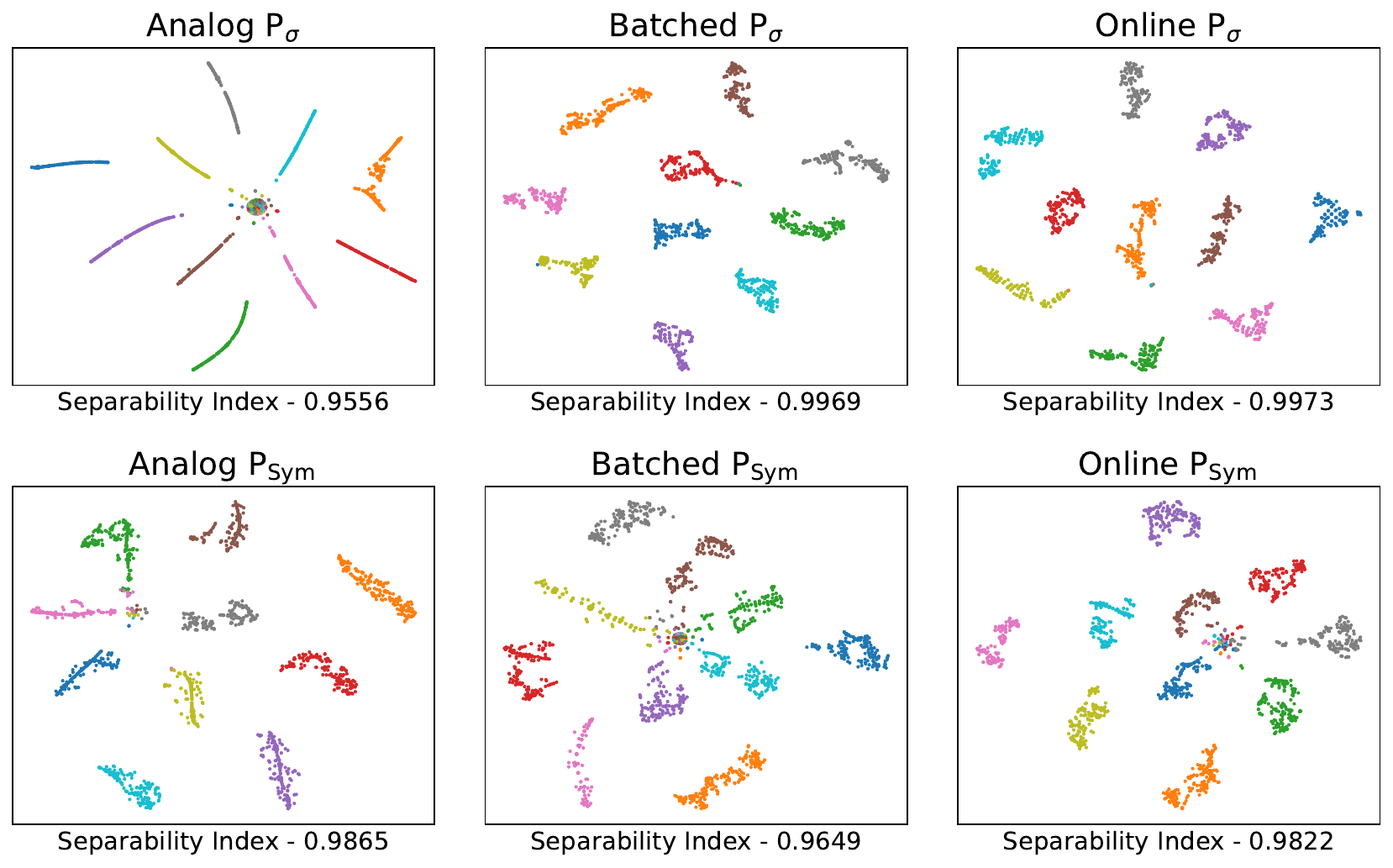}
    \caption{T-SNE projection of the latent space obtained on the test MNIST dataset by all the trained models considered in the experiments for RQ1. Each point represents a projected latent vector, whereas each color represents a different class label. The Separability Index is detailed below each nested plot.}
    \label{fig:latents}
    \vspace{-2mm}
\end{figure}

\paragraph{RQ2:} To answer this second research question, we depict a random sample of latent activity vectors for each model in Figure \ref{fig:latents}, along with their respective Hoyer Index value. The obtained results show that all models achieved sparse latent representations, reaching Hoyer scores above $0.95$ in all cases. Among them, the case of the analog FFA with $P_{\sigma}$ scores best in terms of sparsity, achieving latent distributions where less than $10$ neurons are concurrently active per instance. In contrast, using this probability function on spiking models result in denser yet still sparse latent spaces. This result is probably a byproduct of the additional noise intrinsic to the data preparation process of spiking neural networks. Moreover, when looking into the models with the symmetric probability function $P_{\textup{sym}}$, a converse effect emerges: Hebbian models reach similar sparsity scores as their analog counterpart. Overall, all FFA models can be observed to achieve the same sparse dynamics and neural concentration, where only a tiny set of neurons remain active during the same task.
\begin{figure}[h]
\vspace{-3mm}
    \centering
    \includegraphics[width=\textwidth]{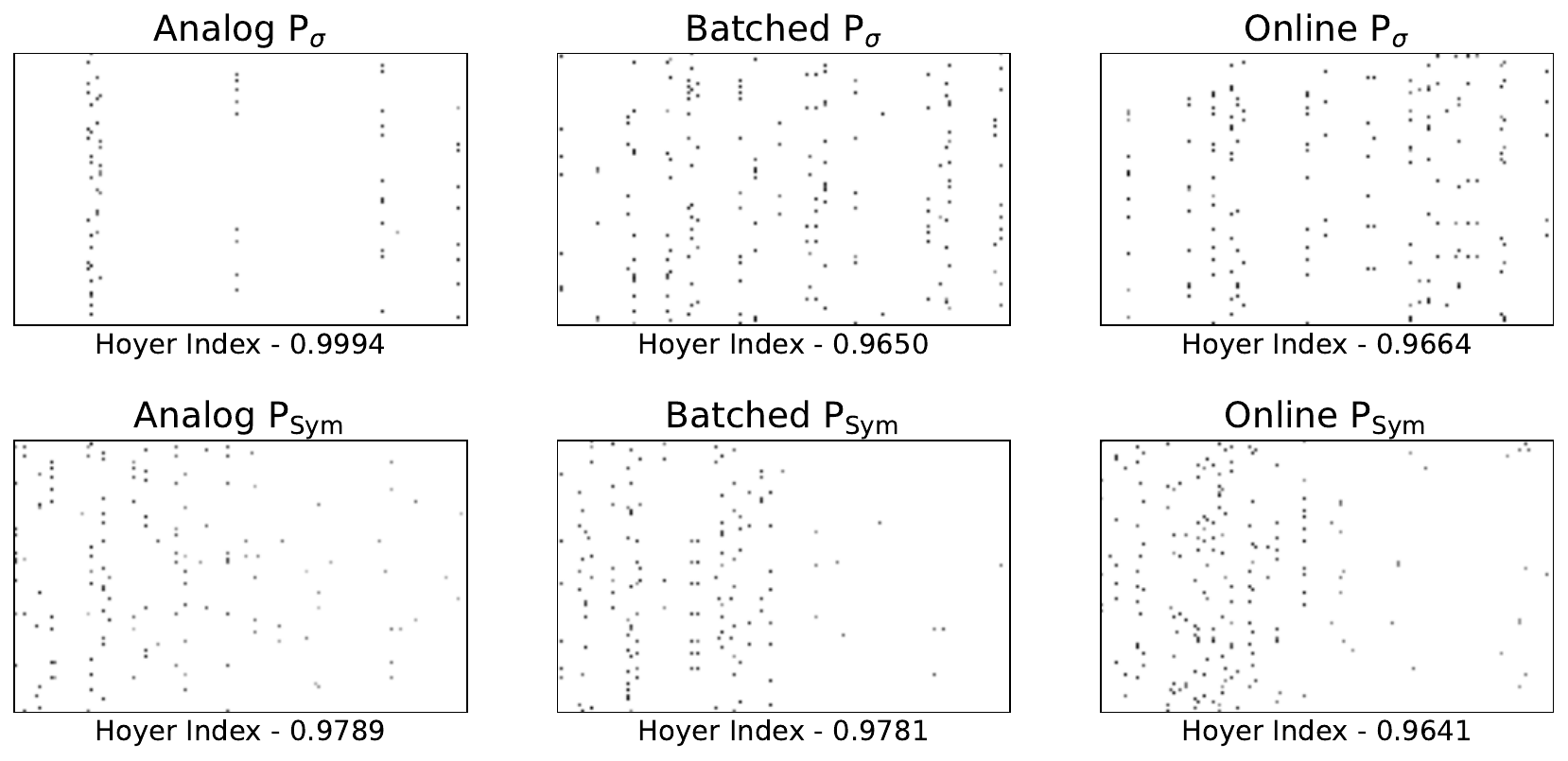}
    \caption{Latent vectors obtained from the test MNIST dataset for all the trained models from RQ1. Each row represents a distinct latent vector. White areas indicate lack of activity, while darker areas indicate greater activity. The Hoyer Index for each model is presented below the latent vectors.}
    \label{fig:tsne_results}
    \vspace{-3mm}
\end{figure}

The 2D projection (via T-SNE \cite{van2008visualizing}) of the latent space for the different models is depicted in Figure \ref{fig:tsne_results}, in which each projected latent vector is colored in terms of the label of its corresponding data instance. Values of the separability index are also indicated under each latent representation. As can be observed in these plots, all models generate highly clustered spaces, where activity vectors of data instances of a given class are projected closely to each other in the latent space. This effect is further verified by the separability index, which shows that samples have an average of at least 95\% nearest neighbors of the same class. Contrary to the previous property, the analog $P_{\sigma}$ results in the least separate space. This effect arises due to a high concentration of samples with activities close to zero, rendering them close within the space. This same effect is observed in all instances of the symmetric probability, which achieve a high separability degree while having clusters of points near the origin. In contrast, the additional noise in the spiking neural networks in the latent spaces of the $P_{\sigma}$ probability allows latent vectors to set themselves away from the origin, fixing the aforementioned effect of the analog case, and achieving high separability in the latent space.

Despite the cluster of points at the origin in the analog $P_{\sigma}$, all model heuristics appear to result in equivalent latent spaces. The most significant difference can be attributed to the reduced dimension of the latent clusters in the analog case. This can be observed in Figure \ref{fig:latents}, where the reduced number of active neurons creates low-dimensional clusters, resulting in curve-like structures instead of point cloud-like structures. Similarly, the space corresponding to the cases using the $P_{\text{Sym}}$ probability function exhibit the same difference between analog and spiking models, where the increased noise from the spiking networks creates less thread-like geometries. However, under this probability function, a small set of latent vectors can be seen to obtain reduced activity values across all models. The conclusion of this analysis indicates that the representational properties of FFA remain consistent across different learning heuristics.

\section{Implications for Biological Learning and Neuromorphic Systems}
\label{sec:impact}

As the need for high-performance computers grows, developing solutions that operate under high throughput and low energy constraints has become an essential objective of modern AI. Neuromorphic systems offer an alternative to conventional GPUs and TPUs, enhancing these characteristics by mimicking the behavior of neurons in the brain. This replaces a central processing approach with a parallel, event-driven mechanism. Consequently, these high parallelization capabilities allow models to achieve faster inference speeds while drastically reducing energy consumption.

With the goal of developing a biological alternative to BP, Hinton's FFA replaces the global backward path with local learning rules. With this approach, FFA overcomes various limitations that make BP unsuitable for neuromorphic chips, such as the weight transport and update lock problems. Currently, this algorithm has demonstrated competitive results in terms of accuracy, while simultaneously showing strong similarities to real neural behavior. However, research on FFA has primarily focused on its original formulation, not truly addressing the potential advantages that motivated the algorithm in the first place.

The results from our work prove an existing equivalence relationship between FFA and Hebbian learning, representing a crucial advance in the practical utility and sustainability of neuromorphic FFA implementations. Notably, while the algorithm has achieved accuracy scores nearly comparable to BP in multiple tasks, the primary argument for its use has been its theoretical biological plausibility. However, advancements in its analog forms have not yet provided any competitive advantages that make FFA a preferable option for production-ready models. Nevertheless, this equivalence creates a practical pathway whereby advances in its analog form can benefit its spiking implementations, which can subsequently be leveraged by implementing them in neuromorphic hardware. For instance, the characteristic properties of FFA, namely its sparse latent activity and highly specialized neural representations, are well-suited for event-driven systems where activity is focused only in active areas, thereby allowing FFA to capitalize on lower energy usage. Similarly, the high parallelization properties could be exploited by layer-local losses to enhance training speed by avoiding the update locks required in BP.

To fully leverage the potential of this algorithm, it is crucial to focus on behavioral patterns that resemble those observed in real neural systems. As demonstrated in the previous section, FFA produces sparse latent spaces with a high degree of neural specialization and intra-class latent similarity. This combination is ideal for developing interpretability mechanisms to enhance decision-making in automated systems. Moreover, the intra-class latent similarity provides a robust foundation for improving model accuracy in noisy environments, thereby yielding more reliable models. Algorithms developed using these properties, coupled with the speed advantages of neuromorphic hardware, could lead to fast and robust neural models, which are particularly critical in high-risk scenarios where automated decision has direct consequences to human life.

In addition, our findings presented in this work contribute positively to the broader field of neuroAI, offering an incremental step toward closing the accuracy gap between analog and biological learning approaches. By adapting FFA into a Hebbian-based rule, we demonstrate the effectiveness of neoHebbian methods in achieving a competitive performance on standard supervised learning tasks. Furthermore, due to the latent-driven regulation of the modulating factor, FFA provides a promising method for achieving stable learning. As we know, achieving stability without vanishing or exploding weights or latent activity remains an elusive task, often requiring carefully crafted homeostatic heuristics \cite{zenke2017temporal,carlson2013biologically}. Due to the bounded and monotonic behavior expected of probability functions in FFA, weight updates can be proven to converge, leading to bounded maximal activities and thereby creating stable learning dynamics, with the modulation acting as a homeostatic regulator.

\section{Conclusion and Future Work} \label{sec:conclusion}

The field of neuromorphic computing has been significantly influenced by advances in both AI and neuroscience. Recently, a surge of biologically inspired algorithms has emerged in the AI community, with the Forward-Forward Algorithm (FFA) standing out as a promising alternative to Backpropagation (BP). FFA utilizes a layer-wise training heuristic motivated by biological constraints. However, limited work has been done to study its biological plausibility. 

In this manuscript we have proven that the use of a squared Euclidean goodness in the FFA results in a weight update that is equivalent to a modulated Hebbian learning rule, which we have denoted as \emph{Hebbian FFA}. To empirically validate this connection, we have conducted experiments to gauge the accuracy and to examine latent representations obtained by Hebbian FFA. Our results have verified that Hebbian FFA not only performs at close accuracy levels to that of its analog version, but also produces similar latent activation spaces in terms of class separability and neural specialization. We have also elaborated on the practical consequences of using Hebbian FFA in conjunction with neuromorphic systems for training spiking neural networks, pinpointing its potential for the explainability, sustainability and robustness of these models when deployed in high-risk open-world scenarios. 

Building on the findings reported in this work, we envision two directions for further research. Firstly, we plan to develop software tools to implement FFA on neuromorphic hardware, facilitating further experimentation and accessibility to the Hebbian-FFA algorithm and other algorithmic variants emerging therefrom. Secondly, we will delve into the geometric properties of the latent space induced by the Hebbian FFA update rule (neural specialization, high separability) to derive mechanisms to issue instance-based explanations based on the features driving the output of the network.

\begin{credits}
\subsubsection{\ackname}
The authors thank the Basque Government for its funding support via the consolidated research groups MATHMODE (ref. T1256-22) and D4K (ref. IT1528-22), and the colaborative ELKARTEK project KK-2023/00012 (BEREZ-IA). E. B. Terres-Escudero is supported by a PIF research fellowship granted by the University of Deusto.

\subsubsection{\discintname}
The authors have no competing interests to declare that are relevant to the content of this article.
\end{credits}

\bibliographystyle{splncs04}
\bibliography{mybibliography}

\appendix 
\section{Output Trace Models}
\label{ap:out_trace}

This appendix outlines the three trace models employed as spiking adaptations of analog activation functions. These traces are used to aggregate the spikes of neurons in order to obtain a smooth approximation of the spiking dynamics of the neuron at a given time:
\begin{itemize}[leftmargin=*]
\item \textbf{LI Trace}: This output trace employs a leaky integrator to capture the incoming activity \(\textup{I}_t\), decaying the activity by a factor of \(\tau_{o}\) at each timestep. The trace $T_t$ at a timestep $t$ is given by:
\begin{equation}
    \textup{T}_{t+1} = \mu \textup{I}_t + \tau_{o} \textup{T}_t,
\end{equation}
where $\mu$ denotes the increased step value, $I_t$ represents the spike input, and $\tau_o$ denotes the decay of the trace.
\vspace{3mm}

\item \textbf{Hard-LI Trace}: This trace model, originally introduced in \cite{ororbia2023contrastive}, replaces small increments in activity with an abrupt change that sets the trace to a maximal value. The trace is defined as:
\begin{equation}
    \textup{T}_{t+1} = \textup{I}_t + \tau_{o} (1-\textup{I}_t) \textup{T}_t,
\end{equation}
where $I_t$ represents the spike input and $\tau_o$ denotes the decay of the trace.
\vspace{3mm}

\item \textbf{ReLU Trace}: In contrast to the previous two models, where a constant decay regulated the activity of the neuron, the ReLU trace aggregates all incoming activity, mimicking the behavior of a ReLU function. This trace was employed to investigate whether using a method with the same derivative as ReLU enhances accuracy by providing more precise modulatory activity. The trace is given by:
\begin{equation}
    \textup{T}_{t+1} = \mu \textup{I}_t + \textup{T}_t,
\end{equation}
where $\mu$ denotes the increased step value and $I_t$ represents the spike input.
\end{itemize}
\end{document}